%% file: neurips_2019.tex
\documentclass{article}

     \usepackage[preprint,nonatbib]{neurips_2019}

\usepackage[utf8]{inputenc} 
\usepackage[T1]{fontenc}    
\usepackage{hyperref}       
\usepackage{url}            
\usepackage{booktabs}       
\usepackage{amsfonts}       
\usepackage{nicefrac}       
\usepackage{microtype}      
\usepackage[mode=text]{siunitx} 

\usepackage[ruled,vlined,linesnumbered]{algorithm2e}
\usepackage{subfigure}
\usepackage{color,xcolor}
\usepackage{graphicx}
\usepackage[toc]{appendix}
\usepackage{amsmath}

\usepackage{amsmath}
\DeclareMathOperator*{\argmax}{arg\,max}

\usepackage[colorinlistoftodos,textsize=tiny,textwidth=20mm]{todonotes}

\title{Unsupervised Object Segmentation with Explicit Localization Module}

\author{%
Weitang Liu\\
Department of Electrical Engineering\\
University of California, Davis\\
Davis, CA 95616, USA \\
\texttt{wetliu@ucdavis.edu} \\
\And
Lifeng Wei\\
Department of Statistics\\
University of California, Davis\\
Davis, CA 95616, USA \\
\texttt{lfwei@ucdavis.edu} \\
\And
James Sharpnack\\
Department of Statistics\\
University of California, Davis\\
Davis, CA 95616, USA \\
\texttt{jsharpna@ucdavis.edu}
\And
John D. Owens\\
Department of Electrical Engineering\\
University of California, Davis\\
Davis, CA 95616, USA \\
\texttt{jowens@ucdavis.edu}
}

\begin{document}

\maketitle

\input{tex/Abstract}
\input{tex/Introduction1}
\input{tex/Models}
\input{tex/RelatedWork}
\input{tex/Experiment}
\input{tex/Discussion}

\input{neurips_2019.bbl}
\appendix
\input{tex/Appendix}

\end{document}

%% file: tex/Abstract.tex
\begin{abstract}
In this paper, we propose a novel architecture that iteratively discovers and segments out the objects of a scene based on the image reconstruction quality. Different from other approaches, our model uses an explicit localization module that localizes objects of the scene based on the pixel-level reconstruction qualities at each iteration, where simpler objects tend to be reconstructed better at earlier iterations and thus are segmented out first. We show that our localization module improves the quality of the segmentation, especially on a challenging background. 
\end{abstract}

%% file: tex/Introduction1.tex
\section{Introduction}

A crucial part of human intelligence is scene understanding, which means decomposing the scene into objects and discovering their relationships. Here we mainly focus on object segmentation, an important method for scene decomposition. In a supervised learning scheme, recent methods~\cite{Long_2015_CVPR,Novotny_2018_ECCV} mainly rely on convolution neural network (CNN) or its variant to minimize the deviance between generated object masks and the ground truth. In an unsupervised learning scheme, traditional pixel clustering methods~\cite{delong2008scalable} lead to more sophisticated image clustering methods and loss with CNN~\cite{kkanezaki2018_unsupervised_segmentation}. Other models achieve state-of-the-art performances by learning object masks and imagine reconstruction~\cite{burgess2019monet} through one network~\cite{IODINE} or two separate networks. Inspired by the observation that typical unsupervised models learn to reconstruct relatively simple backgrounds in early epochs and then more complicated details in later epochs, we believe that different parts of images have a different level of reconstruction difficulty. We argue that those higher-level details are the objects, which usually should not share characters with background and thus harder to be reconstructed during the first few epochs.

We approach this problem from pixel-level clustering and iterative object segmentation, similar to MONet~\cite{burgess2019monet}. We propose a network that removes MONet's attention network and segments objects one-by-one through the reconstruction quality mask. Since MONet's attention network does not rely on a reconstruction image before updating the network parameters, which may lead to inconsistencies between the reconstruction image and attention mask, our model utilizes a reconstruction image that decides which area to focus, which instead leads to a consistent approach that ``where the network reconstructs is where it focuses.'' In our method, the first step segments the background and then later steps ``fill in'' the image details missing from the ``first impression.'' Objects are considered higher-level details of the scenes that cannot be easily reconstructed by the ``first impression'' of the scene , i.e. the background. Moreover, a group of pixels should be considered as an object only if its parts move as a whole (except for deformative objects). Thus, objects should be considered through a clear and explicit localization mask.

Our contributions are as follows. We propose a new algorithm that localizes the areas needed for object segmentation by directly measuring reconstruction quality. This coarse-grained estimate of the focused area is then fine-tuned by a Gaussian Mixture Model with very few components that cluster the pixels in that area to obtain a detailed boundary for the object. More importantly, compared with models that rely on network output attention masks, our model has an explicit localization module that guarantees masks of objects are localized. We show that this explicit localization module is necessary for more complicated datasets, such as Montezuma's Revenge, where some objects share similar color but have different modeling difficulties.

%% file: tex/Models.tex
\section{Models}\label{sec:models}
Review of MONet is in Section~\ref{subsec:monet} and our motivation of model design is in Section~\ref{subsec:motivation}. Section~\ref{subsec:quality} discusses how we measure the quality of reconstruction and Section~\ref{subsec:GMM} explains how we conduct the local clustering through GMM. We observe that the reconstruction quality between the background and the objects are adversarial with each other, so we propose a method that mitigates this effect in Section~\ref{subsec:adversary}. The overview of our model and algorithm is Section~\ref{subsec:whole}.

\subsection{MONet Overview}\label{subsec:monet}
As a recent work of unsupervised scene decomposition, MONet is an important framework on which our model is based. The main idea of MONet is to learn to reconstruct the input image by identifying one object at a time. To achieve this goal, an attention model is trained and outputs an attention mask, which claims to focus on one single object in a given image. A VAE is trained to reconstruct the object covered by this attention mask. Furthermore, VAE also tries to recover the attention mask obtained by the attention model to stabilize the training. The network is tuned so that the ``background'' is always given attention in the first step.

To be specific, if we use $\mathbf{s}^{(k)}$ to represent the ``unexplained ratio'' of each pixel after the $k$-th step, the ``explained ratio'' at $k$-th step as $\mathbf{m}^{(k)}$, and the original image input as  $\mathbf{x}$, the idea of MONet can be written as:
\begin{align*}{}
    \mathbf{s}^{(0)} &= \mathbf{1}&\\
    \mathbf{s}^{(k)}&=\mathbf{s}^{(k-1)}(1-\alpha_\phi(\mathbf{x};\mathbf{s}^{(k-1)}))&\quad 1\le k<K\\
    \mathbf{m}^{(k)}&=\mathbf{s}^{(k-1)}\alpha_\phi(\mathbf{x};\mathbf{s}^{(k-1)})& \quad 1\le k<K\\
    \mathbf{m}^{(K)} &= 1-\sum_{k=1}^{K-1} \mathbf{m}^{(k)}&
\end{align*}
where the $\alpha_\phi$ is a trainable attention network parameterized by $\mathbf{\phi}$ and $K$ is the total number of segmentation steps. We adopt this framework but use a different method, which we detail in later sub-sections, to find the mask at each step.

\subsection{Model Motivation}\label{subsec:motivation}
The idea of MONet, which is basically to use attention as masks and cover the image step by step, is natural. But in practice, we found that it is very hard to tune the hyperparameters to learn a good attention model. Specifically, it's hard to make sure that the CNN-based attention model finds masks for objects. This motivates us to propose a new method with an explicit nonparametric localization module that helps find objects. To produce better object boundaries, we considered simpler clustering methods like GMM or KNN and used traditional features like color or location. Besides, instead of finding an accurate attention mask for the objects directly, we choose to find a larger area that contains the object inside and split the object from the local area in the second step. The smaller area makes it possible to use a simple clustering algorithm to detect the object in different remaining not-yet-covered parts.

\subsection{Reconstruction Quality Estimates and Coarse Grained Attention}
\label{subsec:quality}
 We first measure the quality of the reconstruction by the pixel-wise square error between the reconstruction and the input images. The pixel-wise reconstruction quality, $Q_{i,j}$, is measured by
\begin{align*}
    Q_{i, j}=\exp(-\frac{\mathbf{s}||\mathbf{x}_{i, j} - \mathbf{x}^{re}_{i, j}||^2}{2\sigma^2}),
\end{align*}
where $\mathbf{x}^{re}$ is the reconstruction image and $\mathbf{x}$ is the original image. $\mathbf{x}_{i, j}$ and $\mathbf{x}^{re}_{i, j}$ are RGB vectors at the corresponding pixels determined by $\{i, j\}$. $\sigma$ is a hyperparameter for variance.
In order to locate an object roughly by reconstruction quality, both the pixel reconstruction and its neighbor pixels' reconstruction matter. To evaluate that, we used a non-trainable convolution kernel with $\textit{weights}=1$, $\textit{stride}=1$, and SAME padding on $\mathbf{Q}$. Relying on non-trainable kernels, we can avoid trivial solutions found by neural networks. The output of this convolution indicates how well the network reconstructs at each local region; partially overlapping regions are allowed. We then find the location with the highest output value, denoted by $(i_c, j_c)$. This pair of coordinates indicates the center of the region with the best reconstruction quality, and our model generates a rough attention $\mathbf{G}$ based on $(i_c, j_c)$. Given our rough assumption that attention has peak 1 in the center and decays gradually, we choose a Butterworth filter~\cite{ctx3916707340003821} to model attention in a region. Mathematically, a Butterworth filter is
\[G(r, n, f) = \frac{1}{\sqrt{1+(\frac{r}{f})^{2n}}}.\]
Here $r$ is the distance from a point to the center and $n, f$ are hyperparameters. With the center point determined by $\mathbf{Q}$, we can easily obtain the local mask by this filter.

In practice we treated the horizontal and veritcal coordinates independently and the Butterworth filter attention $\mathbf{G}$ on pixel $(i, j)$ is
\[\mathbf{G} = G(|i-i_c|, n, f)\cdot G(|j-j_c|, n, f).\]

\subsection{GMM}\label{subsec:GMM}
A Gaussian mixture model (GMM) is capable of finding components when we have samples from a mixture of separate Gaussian distributions. In this segmentation task, it is possible to approximate the distribution of pixels if we treat them as 5-d Gaussian random variables. The 5 dimensions are RGB channels and two coordinates, denoted as
\[\mathbf{y}_{i, j} = (\mathbf{x}_{i, j}, i, j).\]

For a general GMM, if we know there are $k$ groups and we somehow initialized their means as $\mathbf{\mu}^{(1)}, \mathbf{\mu}^{(2)}, \cdots, \mathbf{\mu}^{(k)}$ and let
\[z_{i,j,k} = P(\mathbf{y}_{i, j} \textnormal{ in class }k),\quad \mathbf{z}_i\in\mathbb{R}^k,\]

we can easily obtain the update rule for GMM:
\begin{align*}
    \mathbf{\mu}^{(k)} &= \frac{\sum_{i, j} z_{i,j, k}\mathbf{y}_{i, j}}{\sum_{i, j} z_{i,j,k}}\quad k\in\{1,2,\dots\}\\
    z_{i,j,k} &= \frac{\mathcal{K}(\mathbf{y}_{i, j}, \mathbf{\mu}^{(k)})}{\sum_k \mathcal{K}(\mathbf{y}_{i, j}, \mathbf{\mu}^{(k)})}
\end{align*}

Here $\mathcal{K}(\cdot, \cdot)$ is a kernel function. In GMM, since we assume each group is a Gaussian distribution, we choose the multi-variate Gaussian density function as our kernel.

For our problem, the pixels in which we are interested might have been partially/totally explained by previous steps. So they should have less/no impact on the clustering process in later steps. Thus we give each pixel a ``weight'' that indicates its importance. The weight is easy to find: simply use the Butterworth filter weight $\mathbf{G}^{(k)}$:
\begin{align*}
    \mathbf{\mu}^{(k)} &= \frac{\sum_{i, j} z_{i,j, k}w_{i, j}\mathbf{y}_{i, j}}{\sum_{i, j} z_{i,j,k}}w_{i, j}\quad k\in\{1,2,\dots\}\\
    z_{i,j,k} &= \frac{\mathcal{K}(\mathbf{y}_{i, j}, \mathbf{\mu}^{(k)})}{\sum_k \mathcal{K}(\mathbf{y}_{i, j}, \mathbf{\mu}^{(k)})}
\end{align*}


And in practice we found that giving a hard threshold gives clearer segmentation, so we use
\[\mathbf{w} = \mathbf{1}_{\mathbf{G}^{(k)} > 0.5}\]

\begin{algorithm}
\SetAlgoLined
\KwResult{Image reconstruction and segmentation masks}
 Initialize $\mathbf{z}_{i, j, k}=0.5$ for $k=1, 2$;\\
 Initialize $\mathbf{\mu_1}=\mathbf{0}, \mathbf{\mu_2}\sim \textit{Unif}(0, 1)$;\\
 Input vectors $\mathbf{y}_{i, j}$;\\
 Calculate $\mathbf{G}^{(k)}$ based on the Butterworth filter;\\
 Initialize weights by $\mathbf{w}=\mathbf{1}$ if $\mathbf{G}^{(k)} > 0.5$;\\
 \For{L iterations} {
    Update $\mathbf{\mu}^{(k)} = \frac{\sum_{i, j} z_{i,j,k}w_{i,j}\mathbf{y}_{i,j}}{\sum_{i,j} z_{i,k}w_{i,j}}, k=1, 2$;\\
    Update variance for each coordinate for each group;\\
    Update $z_{i,j,k}= \frac{\mathcal{K}(\mathbf{y}_{i,j}, \mathbf{\mu}^{(k)})}{\sum_{k=1}^2 \mathcal{K}(\mathbf{y}_{i, j}, \mathbf{\mu}^{(k)})}$
 }
 $I_{i, j}= z_{i, j, 2}$;\\
 return $\mathbf{I}$
 \caption{$L_{GMM}$\label{alg:2}}
\end{algorithm}

\subsection{Adversary Between background and object reconstruction}\label{subsec:adversary}

A problem we observe while using reconstruction quality as masks is that the more details the background could obtain through training, the worse the segmentation result will be, because it leaves less room for perfect reconstruction of objects in later iterations. In order to mitigate this problem, we generate a mask for the background by subtracting all the intermediate objects' attention masks. The portion of the image that should be captured by the background at the first iteration is derived from the input image masked by this background mask. In other words, $\mathbf{m}^{(1)}$ for background is computed by
\begin{align*}
    \mathbf{m}^{(1)} = 1-\sum_{k=2}^{K} \mathbf{m}^{(k)}.
\end{align*}

\subsection{Model Overview}\label{subsec:whole}
Our model utilizes VAE~\cite{kingma2013auto} for simple datasets or an auto-encoder with skip-connection for complicated datasets. We assume there are $K$ objects, including background and our model keeps track of the not-yet-explained areas by a remaining-mask $\mathbf{s} \in[0,1]^{h\times w}$. As an alternative, we can use stick-breaking process to find just the right number of K. In every iteration, our model tries to reconstruct the unexplained part. We evaluate the reconstruction quality and denote the quality by $\mathbf{Q} \in[0,1]^{h\times w}$. A basic observation is that VAEs and auto-encoders will learn the representation of often observed items so the reconstruction quality of these objects will get better faster than other areas. Thus, except for the first iteration where the background is found, we can locate the object roughly by finding the area where reconstruction quality is high. Then we look into the area and tell whether each pixel belongs to the object. As for the first iteration, we directly use the quality of reconstruction of each pixel as its explained ratio by the background.

For the following iterations, the image- and remaining-mask are input to the network again for reconstruction. Different from the first iteration, a location-sensitive mask, denoted by $\mathbf{L} \in[0,1]$, is derived from the location where the reconstruction quality is the best. Then the remaining-mask is updated again and the next iteration starts. The remaining-mask $\mathbf{s}$ and object component mask $\mathbf{m} \in[0,1]$ are updated by the formula:
\begin{align*}
    \mathbf{s}^{(k)}&=\mathbf{s}^{(k-1)}(1-\mathbf{Q}^{(k)}\mathbf{L}^{(k)})\\
    \mathbf{m}^{(k)}&=\mathbf{s}^{(k-1)}\mathbf{Q}^{(k)}\mathbf{L}^{(k)},
\end{align*}
where $k \in\{1,2,\cdots,K\}$ denotes the $k$th iteration. Moreover, when $k$ is 0, $\mathbf{L}_1$ is $\mathbf{1}$ to indicate no location-sensitive mask is applied to the background.

\begin{figure}[h]
\begin{center}
\includegraphics[width=0.85\textwidth]{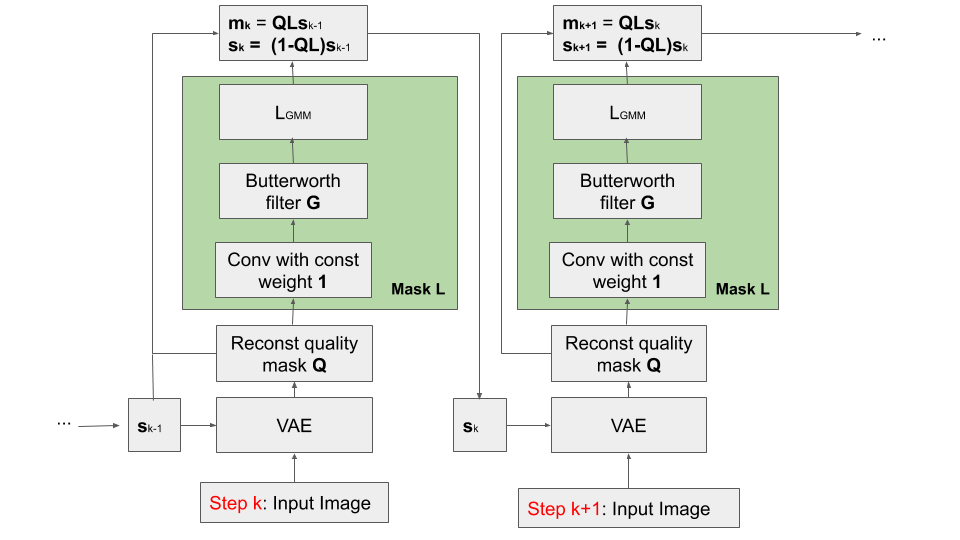}
\end{center}
\caption{Our model structure and how model $\mathbf{m}$ and $\mathbf{s}$ are computed.\label{fig:flow_chart}}
\end{figure}

Denote $c$ as input image's channel, $\theta$ as the trainable weights for encoder-decoder, $\phi$ as the constant weights for a convolution layer that evaluates the area reconstruction quality from pixel-wise reconstruction quality mask $\mathbf{Q}$, $\sigma_1$ as the constant variance for pixel-wise reconstruction quality mask, and $\sigma_2$ as the constant variance for location mask $\mathbf{L}$. The algorithm we use is summarized in Algorithm~\ref{alg:1} and its flowchart is in Figure~\ref{fig:flow_chart}.


\begin{algorithm}
\SetAlgoLined
\KwResult{Image reconstruction and segmentation masks}
 Initialize a remaining-mask $\mathbf{s}^{(0)}$ as $\mathbf{1}$.\;
 \For{K iterations} {
  $\mathbf{x}^{re,(k)} = g_\theta(\mathbf{x}, \mathbf{s}^{(k-1)})$ \tcp*{decode}\
  $\mathbf{Q}^{(k)} = \exp(-\frac{\mathbf{s}^{(k-1)}\sum_c(\mathbf{x}^{re,(k)}_c-\mathbf{x_c})^2}{2\sigma_1}) \quad c \in\{0,1,2\}$\tcp*{decode quality mask}\
  with no-gradient:\\
  \Indp
   $\mathbf{Q'}^{(k)} = Q_{\phi=\mathbf{1}}(\mathbf{s}^{(k-1)}\mathbf{Q}^{(k)})$\tcp*{area decode quality}\
   $i,j = \argmax(\mathbf{Q}^{(k)})$\;
   $\mathbf{L}^{(k)} = L_{GMM}(\mathbf{x}, i, j; \sigma_2 )$\tcp*{location mask~\ref{alg:2}}\
   $\mathbf{s}^{(k)}=\mathbf{s}^{(k-1)}(1-\mathbf{Q}^{(k)}\mathbf{L}^{(k)})$\;
   $\mathbf{m}^{(k)}=\mathbf{s}^{(k-1)}\mathbf{Q}^{(k)}\mathbf{L}^{(k)}$\;
 }
 \caption{Background and Object Segmentation\label{alg:1}}
\end{algorithm}

The only trainable parameter $\theta$ is updated through the following equation:
\begin{equation}\label{equ:loss}
\begin{split}
    \mathbf{Loss}=&\sum^{K}_{k=1}\mathbf{m}^{(k)}(\mathbf{x}^{\textit{re},(k)}-\mathbf{x}^{\textit{gt},(k)})^2\\
    & + \beta\sum^{K}_{k=1}(\mathbf{1}-\mathbf{m}^{(k)})(\mathbf{x}^{\textit{re},(k)}-\zeta)^2 \\
    &+ \gamma \sum^{K}_{k=1}D_{KL}(p(\mathbf{\tilde{z}}^{(k)}|\mathbf{x}, \mathbf{s}^{(k-1)})||p(\mathbf{\tilde{z}}))
\end{split}
\end{equation}

where $\zeta$ is a constant prior for pixels that are masked out by $\mathbf{m}$, $\beta$ is a hyperparameter that controls the weight of prior loss, and $\gamma$ is the hyperparameter that controls the weight of KL-div for the VAE prior. For auto-encoder $\gamma=0$, $\mathbf{\tilde{z}}$ is the prior for enbedding of VAE and $\mathbf{\tilde{z}}^{(k)}$ is the embedding in the $k$th step. Lastly, if we use $\mathbf{m}^{(k)}$ as the object mask, then redefining $\mathbf{x}^{re,(k)}:=\mathbf{m}^{(k)}\mathbf{x}^{re,(k)}$ for Equ.~\ref{equ:loss} helps, since the reconstruction at this time focuses more  on the region found by $\mathbf{m}^{(k)}$. This allows mask $\mathbf{Q}$ to contribute to the loss, but $\mathbf{L}$ is strictly constant.

%% file: tex/RelatedWork.tex
\section{Related Work}
\label{sec:related}

A lot of the recent progress has been made as a result of convolutional neural networks (CNN) and their variants. Many of the progress on object segmentation has been based on supervised learning, where people label the images given their prior knowledge and train the network accordingly. Fully Convolutional Networks~\cite{Long_2015_CVPR} is a paradigm network architecture for semantic segmentation and more advanced results are achieved by recently with a semi-convolutional operator~\cite{Novotny_2018_ECCV}.

People also work on unsupervised object segmentation through neural network. Unsupervised object discovery~\cite{hsu2018co} with a pre-trained model, such as VGG, is proposed, but it is highly based on the performance of the pre-trained model; the model is not end-to-end unsupervised. An object segmentation method characterizes pixel similarities based on CNN. Recently, a generative adversarial network (GAN) for object segmentation~\cite{chen2019unsupervised} is successfully applied to real world dataset. However, this model is based on the network itself to find attention masks, which are good in datasets where the objects are salient and big enough.

The most related work is MONet~\cite{burgess2019monet}, which segments objects iteratively through a scope mask. Its performance relies on the interactions between the attention mask and VAE for reconstruction. A similar idea is also used in IODINE~\cite{IODINE}, where each independent embedding tries to recover an object and its mask for the object. Then all recovered objects are combined with a normalized mask to reconstruct the original image. It uses special techniques to learn the joint posterior embedding to overcome the shortage of VAE, which is only able to learn independent posterior embedding given the input. In practice, tuning the parameters such that the CNN-based attention model masks exactly over objects is very difficult: there is no feedback loop between the reconstruction image and the attention module for every scene-decomposing step in MONet. Therefore, what to decomposed in every step is purely determined by the attention model; it is difficult to guarantee that the attention model masks over a localized region and that region happens to contain an object. Compared to MONet, our model directly computes an attention mask from reconstructed images, and we have an explicit localized module that makes sure our model focuses on local regions.

Lastly, in terms of object discovery through a sequence of frames, a network~\cite{pathak2017learning} that is based on optical flow can learn moving objects. Neural Expectation Maximization~\cite{NEM} proposes to learn embeddings of objects through a sequence of frames through EM and learn the transformation from frames to embeddings through training. Based on NEM, Relational NEM~\cite{van2018relational} where object relationships are extracted through the embedding and R-NEM achieves better performance than NEM\@. Object discovery through a sequence of frames provides more information than independent frames. Since our model currently only focuses on object segmentation on images, this direction is interesting future work.

%% file: tex/Experiment.tex
\section{Experiments}
We test our model on three different datasets. We mainly focus on the performance on object segmentation.

\textbf{Multi-dSprites}~\cite{multiobjectdatasets19} This dataset has a colored background and a random number of objects. We use \num{60000} training samples and \num{10000} testing samples. We use the ARI score provided. We conduct an ablation study on this dataset.

\textbf{The Category Flower Dataset}~\cite{Nilsback08} This dataset is a real-world flower dataset. We use the same data split as provided. We use the sumScoreIoU~\cite{chen2019unsupervised}.

\textbf{Montezuma's Revenge} Montezuma's Revenge OpenAI Gym~\cite{1606.01540} is a game where object discovery plays an important role in hierarchical deep reinforcement learning proposed~\cite{kulkarni2016hierarchical} and goal-driven/symbolic planning for reinforcement learning~\cite{lyu2019sdrl}. We use \num{10000} training samples with a random policy. For testing sets, we use a pre-trained policy where the agent successfully solve the first stage. We manually label 100 samples of nine objects, including the agent, the skull, the rope, the key (may be missing as the agent obtains it), two doors, and three ladders. Since current GAN-based segmentation~\cite{chen2019unsupervised} supports segmenting simple scene (one foreground object only), and official implementations of MONet or IODINE are not available online, we only report our AMI score on Montezuma's Revenge as a benchmark result. The AMI score is calculated as in NEM\@~\cite{NEM}.

\begin{figure}[h]
\begin{center}
\subfigure[]{\includegraphics[width=13mm]{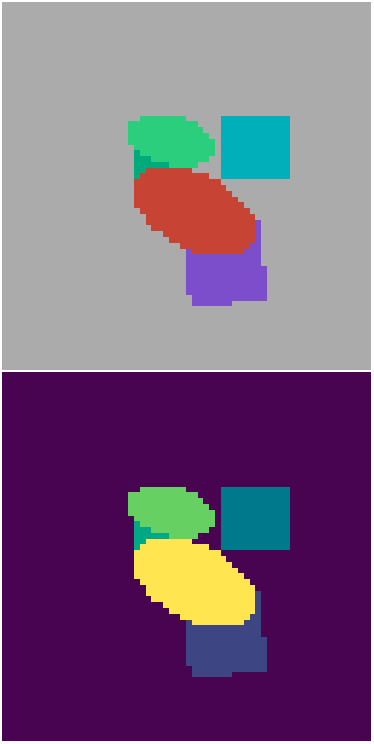}}
\subfigure[]{\includegraphics[width=13mm]{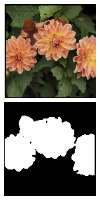}}
\subfigure[]{\includegraphics[width=13mm]{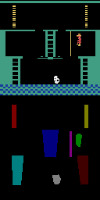}}
\caption{Sample ground truth images (top) and segmentation masks (bottom) for Multi-dSprites (left), Category Flower Dataset (middle), and Montezuma's Revenge (right).}
\end{center}
\end{figure}

\subsection{Results and Discussion}
\begin{table}[t]
\caption{Benchmark results for different dataset and models. The higher the better for all measurements.\label{table:res}}
\begin{center}
\begin{tabular}{llll}
\multicolumn{1}{c}{\bf Dataset}  &\multicolumn{1}{c}{\bf Multi-dSprites}
&\multicolumn{1}{c}{\bf Category Flower Dataset}
&\multicolumn{1}{c}{\bf Montezuma's Revenge}
\\ \hline \\
MONet          & 0.904$\pm$0.008   & ---             & --- \\
IODINE         & 0.767$\pm$0.056   & ---             & --- \\
ReDO           & ---               & 0.764$\pm$0.012 & --- \\
Ours           & 0.621$\pm$0.004   & 0.632$\pm$0.001 & 0.375$\pm$0.002\\
\end{tabular}
\end{center}
\end{table}

We summarize our results in Table~\ref{table:res}. We also analyze the results by different datasets in the following paragraphs.

\textbf{Multi-dSprites} Our model so far is not able to achieve as good results as MONet, because this dataset has samples in which objects are partially covered by another object, a situation that GMM cannot handle easily. For ablation study, where the whole location mask $\mathbf{L}$ is removed and only the reconstruction quality $\mathbf{Q}$ remains, we found that the segmentation metric provided~\cite{multiobjectdatasets19} does not apply to our case, because images with one object leads to NAN as our model incorrectly decomposes that object to different object slots. We see similar poor performance when we leave the GMM module and $\mathbf{L}$ only without training the network.

\textbf{Category Flower Dataset} This is a real world dataset where the background is more complicated and thus requires more sophisticated background removal techniques. Similar to the results achieved when IODONE~\cite{IODINE} applies their model to a real-world dataset, we observe a noticeable gap between our model and the benchmark, because the assumption of 5D vector in GMM module, RGB and x-y coordinates, is too simple for real world dataset.

\textbf{Montezuma's Revenge} Our model can achieve an AMI score of 0.375. Figure~\ref{fig:mrl} is a reconstruction of the objects in a typical frame. Figure~\ref{fig:mr} in Appendix~\ref{A:mr} provides more reconstruction results. Most of the important objects that can serve as goals are found by our model, including the three ladders, skull, the doors, and even the keys. More strikingly, our model successfully finds objects, such as the ladders at the bottom, that share the same color as the walls. Since the wall is easier to reconstruct, it is captured by the background, whereas the ladders, with their more complicated details, are left to be captured by later object slots. Through this experiment, we confirm our argument that objects can be extracted iteratively with different reconstruction difficulties, which we believe can be a new method for object discovery.

\subsection{Object Location Extractor}
\begin{figure}
\centering
\subfigure{\includegraphics[width=10mm]{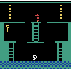}}
\subfigure[]{\includegraphics[width=10mm]{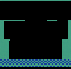}}
\subfigure[]{\includegraphics[width=10mm]{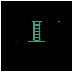}}
\subfigure[]{\includegraphics[width=10mm]{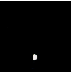}}
\subfigure[]{\includegraphics[width=10mm]{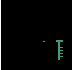}}
\subfigure[]{\includegraphics[width=10mm]{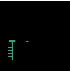}}
\subfigure[]{\includegraphics[width=10mm]{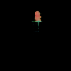}}
\subfigure[]{\includegraphics[width=10mm]{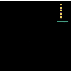}}
\subfigure[]{\includegraphics[width=10mm]{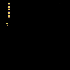}}
\subfigure[]{\includegraphics[width=10mm]{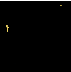}}
\subfigure[]{\includegraphics[width=10mm]{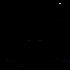}}
\caption{Randomly picked sample of reconstructed image by objects for Montezuma's Revenge, where objects locations are provided. The first figure is input image. Each important object has been found by our model, and its location visually matches with our provided, showing that our GMM module does find the objects.\label{fig:mrl}}
\end{figure}

\begin{table}[h]
  \caption{Location found by objects and their corresponding location.Matching between objects visual position with the x-y coordinates provided shows that our model successfully find the objects.\label{table:mrl}}
\begin{center}
\begin{tabular}{lll}
\multicolumn{1}{c}{\bf Objects}  &\multicolumn{1}{c}{\bf y-axis}
&\multicolumn{1}{c}{\bf x-axis}
\\ \hline \\
back ground         & ---   & --- \\
middle Ladder       & 0.365 & 0.497 \\
skull               & 0.798 & 0.489 \\
bottom right ladder & 0.595 & 0.827 \\
bottom left ladder  & 0.595 & 0.180 \\
agent               & 0.231 & 0.531 \\
right door          & 0.282 & 0.880 \\
left door           & 0.090 & 0.137 \\
key                 & 0.378 & 0.067 \\
---                 & 0.026 & 0.887 \\
\end{tabular}
\end{center}
\end{table}

One of the benefits of our model is that the objects' location is automatically extracted. We train our model in only 9 epochs and extract the locations of the objects through the coordinate means calculated with the GMM\@. A random reconstructed frame is provided as Figure~\ref{fig:mrl} and the corresponding objects and their location is provided in Table~\ref{table:mrl}. The location is shown as scaled from 0--1 in $y$ (vertical axis) and $x$ (horizontal axis). $(0,0)$ is the top left of the image. Most of the objects are found by our model, visually near the locations provided.

%% file: tex/Discussion.tex
\section{Discussion and future work}
In this paper, we propose a new unsupervised object segmentation algorithm with an explicit localization module. The localization module serves as an attention mask derived from the reconstruction quality. By iteratively segmenting the objects, our method finds objects one-by-one, filling in the details of the image missing from previous iterations. We empirically confirm our beliefs that those details correspond to objects.

As for future work, it is promising to extend this work in a sequence of frames, a context where objects are mostly consistent between frames. We also notice that GMM does not always lead to good results and more sophisticated (local) segmentation algorithms could possibly lead to better results. Lastly, our model still has a decent amount of prior knowledge injected through hyperparameters. Making the model simpler should be helpful in future work.

%% file: tex/Appendix.tex
\newpage
\section{Appendix}\label{A:mr}

\begin{figure}[h]
\begin{center}
\includegraphics[width=40mm]{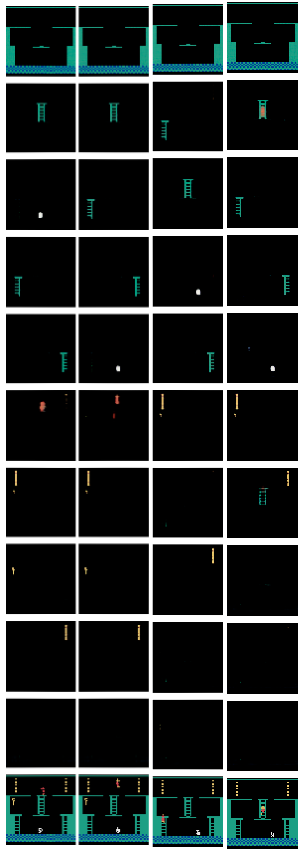}
\end{center}
\caption{Randomly picked samples of reconstructed image by objects for Montezuma's Revenge. Last row is input/ground truth image. These sample reconstruction images demonstrate that our model almost decomposes the scene perfectly into different objects. \label{fig:mr}}
\end{figure}
